\documentclass{article}

\usepackage{nips13submit_e}
\usepackage{times}
\usepackage{epsfig}
\usepackage{graphicx}
\usepackage{amsmath}
\usepackage{amssymb}
\usepackage{multirow}
\usepackage[lined,boxed,commentsnumbered, ruled]{algorithm2e}

\usepackage{hyperref}
\title{Image Representation Learning Using Graph Regularized Auto-Encoders}

\author{Yiyi Liao, Yue Wang, Yong Liu\\
State Key Laboratory of Industrial Control and Technology\\
Zhejiang University, China\\
\texttt{yongliu@iipc.zju.edu.cn}
}

\nipsfinalcopy 

\begin{document}


\maketitle

\begin{abstract}
It is an important task to learn a representation for images which has low dimension and preserve the valuable information in original space. At the perspective of manifold, this is conduct by using a series of local invariant mapping. Inspired by the recent successes of deep architectures, we propose a local invariant deep nonlinear mapping algorithm, called graph regularized auto-encoder (GAE). The local invariant is achieved using a graph regularizer, which preserves the local Euclidean property from original space to the representation space, while the deep nonlinear mapping is based on an unsupervised trained deep auto-encoder. This provides an alternative option to current deep representation learning techniques with its competitive performance compared to these methods, as well as existing local invariant methods.\end{abstract}

\section{Introduction}








%
%
%

Although the dense original image can provide intuitive visual representation, it is well known that this representation may cover the hidden semantic patterns which need to be recognized by those image based learning tasks. On the other side, the performance of machine learning methods is also strongly dependent on the corresponding data representation on which they are applied. Thus image representation becomes a fundamental problem in visual analysis.
Given an image data matrix $X\in\mathbb{R}^{m\times n}$, each column of $X$ corresponding to an image, the image representation is to find a representation function $H=f(X)(H\in \mathbb{R}^{l\times n})$, which can extract useful information from $X$. And each column vector of $H$ is the representation of an image in this concept space.

At the perspective of dimension reduction, the learned representation should have a lower dimension than the original one, i.e. $l<m$, and express the property of data better at the same time. The former is directive, while the later, usually be measured by the performance of clustering $H$. In this paper, we aim on this dimension reduction problem. One of the usual frameworks to model this problem is an optimization problem minimizing a cost shown as
\begin{equation}\label{frameworkdr}
C=\mathit\Phi(X,H)+\mathit\Psi(H)
\end{equation}
where the first term measures the approximation of $H$ to $X$, while the second term, constrains the representation space.

In this paper, we propose an implementation of (\ref{frameworkdr}) based on deep learning and manifold, called graph regularized auto-encoder (GAE). The choice of $\mathit\Phi$ is graph regularizer, which constrains the $H$ to have the similar local geometry of original data space $X$. This is motivated by the property that a manifold resembles Euclidean space near each point (Wiki). Regard $X$ as a manifold, then a neighborhood of each $x$ has Euclidean property, which we want to be kept in $H$. However, whether this can be achieved depends on the choice of $f$, which maps $X$ to $H$. It should have enough expressive power to map the original space to the constrained representation space. So we choose deep network to achieve better performance beyond the existing many interesting linear functions with its nonlinearity. It is also expected that many recent successes on deep learning based approaches in supervised tasks ~\cite{NIPS2012_0534,DBLP:conf/nips/SocherHBMN12} can be extended to the context of unsupervised ones.

The remainder of this paper is organized as follows: In section 2, we will give a brief review of auto-encoder based representation learning and the related works;
Section 3 will introduce our graph regularized auto-encoder for image representation learning tasks including both unsupervised conditions and semi-supervised conditions. Extensive experimental results on clustering are presented in Section 4. Finally, we provide a conclusion and future works in Section 5.

\section{Background}
Auto-Encoder~\cite{HinSal06,DBLP:conf/nips/HintonZ93,10.1109/TPAMI.2013.50} is a special neural network, whose input is same as the output of the network. Given a data set $X=\{x_1,...,x_n\}\in \mathbb{R}^{m \times n}$, each column of $X$ is a sample vector. $H\in \mathbb{R}^{l\times n}$ is a feature representation of the original data set $X$ by an encoder function $H=f_\theta(X)$. Normally, $l<m$, and $H$ can be regarded as a low dimensional representation (or subspace) of the original data set $X$. And another feature mapping function, which is called decoder, maps from feature space back into input space, thus producing a reconstruction $Q=q_\theta(H)$. A reconstruction error function $L(X,Q)$, which is also called loss function, is defined, and the set of parameters $\theta$ of the encoder and decoder are learned simultaneously on the task of reconstructing as well as possible the original input, i.e. attempting to incur the lowest possible reconstruction error of $L(X,Q)$~\footnote{Normally, the loss function is defined as the Euclidian distance of the two data set, that is $\parallel X-Q\parallel^2$. }.

The most commonly used forms for the decoder and encoder are affine mappings, optimally followed by a non-linearity as:
\begin{eqnarray}\label{train}
    H=f_\theta(X)=s_f(b_H+W_HX)\\
    Q=q_\theta(H)=s_q(b_Q+W_QX)
\end{eqnarray}

$s_f$ and $s_q$ are the encoder and decoder activation functions, e.g. non-linear functions of sigmoid and hyperbolic tangent or linear identify function etc. Then the set of parameters is $\theta=\{W_H,b_H,W_Q,b_Q\}$, and the problem is formally presented as follows:
\begin{equation}\label{optimalfunction}
\hat{\theta}=\textrm{arg min}_{\theta} L(X,q_\theta(f_\theta(X)))
\end{equation}

Formula~(\ref{optimalfunction}) can be easily solved by the stochastic gradient descent based backpropagation approaches. The auto-encoders are also able to support multiple layers, e.g in Hinton's work~\cite{HinSal06}, which train the encoder network (from $X$ to $H_i$, $i$ is the number of the layer) one-by-one using the Restricted Boltzamann Machines and the decoder layers of the network are formed by the inverse of the trained encoder layers, such as $W_H=(W_Q)^T$ in one layer auto-encoder.

There are also some regularized auto-encoders such as sparse auto-encoders~\cite{DBLP:conf/nips/RanzatoPCL06,DBLP:conf/nips/LeeEN07,Goodfellow+etal09:invariance,DBLP:conf/icml/LarochelleB08},
denoising auto-encoders~\cite{Vincent:2011:CSM:2000609.2000610,Vincent:2010:SDA:1756006.1953039,Vincent:2008:ECR:1390156.1390294} and contractive auto-encoders~\cite{citeulike:9426230,Rifai:2011:HOC:2034117.2034159}. It is pointed out that the sparse penalty used in sparse auto-encoder will tend to make only few input configurations can have a low reconstruction error~\cite{NIPS2007_1118}, which may hurt the numerical optimization of parameters. The other two kinds of regularized auto-encoders are regarded to make the representation as insensitive as possible with respect to changes in input, which is commonly useful in supervised learning condition, however, it may not provide positive impacts in unsupervised and semi-supervised conditions.

Previous studies have also shown that the locally invariant idea~\cite{Hadsell:2006:DRL:1153171.1153654} will play an important role in the image representation, especially for those tasks of unsupervised learning and semi-supervised learning. There are many successful manifold learning algorithms, such as Locally Linear Embedding (LLE)~\cite{Roweis2000}, ISOMAP~\cite{Tenenbaum2000global}, and Laplacian Eigenmap~\cite{DBLP:conf/nips/BelkinN01}, which all implement the locally invariant idea that nearby points are likely to have similar embeddings. However, these methods are all linear, which may not provide enough expressive power to find a representation space that can preserve the local geometry.

There are some similar works on graph regularized neural network architecture. \cite{mobahi2009deep} proposed a graph regularizer that constrains the similarity between consecutive frames, which shows the human knowledge can be applied in this term. In \cite{min2010deep}, a graph constrains the data points belonging to the same label is proposed. In this work, the deep network is trained, and then minimizes only the graph regularizer using this network. Both these works use the correct graph regularizer since it is built using the correct supervised or human knowledge information.

Another work similar to us is ~\cite{Hadsell:2006:DRL:1153171.1153654}, in which a convolutional neural network is applied to minimize a graph regularizer. In our work, we minimize a combined cost (\ref{frameworkdr}) using a fully connected network. The reason is that, the graph in the unsupervised tasks is not completely correct (pair of data point belonging to the different label actually), an introduction of first term in (\ref{frameworkdr}) can act as a regularizer avoiding fitting of wrong information. Besides, we do not introduce a mechanism that pulls apart discriminative pairs.






\section{Graph Regularized Auto-Encoder}

The problem is to design the second term $\mathit\Psi(H)$ in (\ref{frameworkdr}) to constrain the representation space. In this section, we introduce our geometrical regularization that implement locally invariance in unsupervised and semi-supervised learning.

\subsection{Single Layer Auto-Encoder Regularized with Graph}
In our Graph regularized Auto-Encoder (GAE), both the decoder and the encoder use sigmoid as their activation functions. Denote sigmoid function as $S(x)=1/(1+e^{-x})$. Then the encoder and decoder can be presented as follow:
\begin{eqnarray}\label{dr}
  H = f_\theta(X) &=& S(W_HX+b_H)\\ \label{Hfunction}
  Q = q_\theta(H) &=& S(W_QH+b_Q) \label{Qfunction}
\end{eqnarray}As the representation should discover the latent structure of the original data, the geometrical structure of the data will be the ideal latent structure in representation learning especially in unsupervised or semi-supervised learning. A reasonable assumption is that if two data points $x_i,x_j$ are close in the intrinsic geometry of the data distribution, then their corresponding representations, $h_i,h_j$, should be also close to each other. This assumption is usually referred to as local invariance assumption~\cite{DBLP:conf/nips/BelkinN01,Roweis2000,MR_JMLR_06}, which plays an essential role in designing of veracious algorithms, such as dimensionality reduction and semi-supervised learning.

In manifold learning, the local property of the data space are preserved in the reduced representation. But most algorithms considering this problem is linear. In our GAE, we introduce a locality preserved constraint to the nonlinear auto-encoder to better reflect its nature of manifold. Based on formula~(\ref{optimalfunction})-(\ref{Qfunction}), we optimize the auto-encoder regularized with graph as follows
\begin{equation}\label{gtrain}
    \hat{\theta} = \arg \min (\|X-Q\|^2+\lambda tr(HGH^T))
\end{equation}
where $\lambda$ is a coefficient of the training algorithm, $tr(HGH^T)$ is the term of graph regularizer, $tr(\cdot)$ denotes the trace of a matrix, and $G$ is a graph coding the local property of the original data $X$. Denote $v_{ij}$ as the weight for the locality between data sample $x_i$ and $x_j$, and their corresponding representations are $h_i$ and $h_j$. Based on the local invariance assumption, the regularization that requires the points in subspace keeping the same geometrical structure as the original data can be presented as the following weighted formula.
\begin{equation}\label{G}
\begin{split}
    &\sum_i \sum_j v_{ij} \|h_i-h_j\|^2\\
    =&\sum_i h_i^T \sum_j v_{ij} h_i + \sum_j h_j^T \sum_i v_{ij} h_j - 2\sum_i \sum_j h_i v_{ij} h_j\\
    =&tr(H D_1 H^T) + tr(H D_2 H^T) - 2tr( H V H^T)\\
    =&tr(HGH^T)
\end{split}
\end{equation}
where $v_{ij}$ are entries of $V$, and $V$ is the weight matrix, $D_1$ nd $D_2$ are diagonal form with $D_{1,ii} = \sum_j v_{ij}$ and $D_{2,jj} = \sum_i v_{ij}$, and $G=D_1 + D_2 - 2V$. With this constraint, the local property of the data in original space will be preserved after auto-encoder mapping. The matrix $G$ has significant expressive power of the structure in the original data space. It is calculated from the weight matrix $V$, whose design will be introduced in section~\ref{graphdesignsection}.

Formula~(\ref{gtrain}) can be solved by the Broyden-Fletcher-Goldfarb-Shanno (BFGS) training algorithm. In our GAE, the weight-tying constraint is not used, which means our GAE does not require $W_H = (W_Q)^T$.

\subsection{Multiple-layer Auto-Encoders Regularized with Graph}\label{mlautoencoder}

The auto-encoder was proposed in the context of the neural network, which is later applied to train the deep structure of networks to obtain better expressive power. In data representation, the idea of representing with multiple layers still works. Thus the representation of the original data can be presented with one layer of mapping, as well as multiple-layer mapping. We also implement the locally invariant constraint into the multiple layer auto-encoders by adding the graph regularized terms.

As training all the layers simultaneous in multiple-layer auto-encoders may be stacked, in our multiple-layer GAE, we train the multiple-layer GAE layer-by-layer. We use $H_i$ to denote the data representation of the $i$th layer, and its corresponding decoder is denoted as $Q_i$. The input data of the $i$th layer is the data representation of $i-1$ th layer~\footnote{If $i=1$, then the input data is the original data set $X$, and the training process is back to single layer GAE.}.  That is:
\begin{eqnarray}\label{mdl}
  H_i = f_{\theta_i}(H_{i-1}) = S(W_{H_i}H_{i-1}+b_{H_i})\\
  Q_i = q_{\theta_i}(H_i) = S(W_{Q_i}H_i+b_{Q_i})
\end{eqnarray}

Here, $\theta_i=\{W_{H_i},W_{Q_i},b_{H_i},b_{Q_i}\}$, and then the objective function of the $i$th layer of the GAE is,

\begin{equation}\label{gmltrain}
    \hat{\theta_i} = \arg \min (\|H_{i-1}-Q_i\|^2+\lambda tr(H_{i}G_{i-1}H_{i}^T))
\end{equation}

Where $G_{i-1}$ is the graph regularizer generated from data $H_{i-1}$.

Then the Multiple-Layer GAE (ML-GAE) algorithm can be given as follow:

\begin{algorithm}
\caption{ML-GAE}
\SetKwData{Index}{Index}

\SetAlgoNoLine\LinesNumbered \KwIn{$X$, total layer number $j$}
\KwOut{$W_{H_1},b_{H_1},...,W_{H_j},b_{H_j}$ }
\BlankLine

\For{ $i=1$ to $j$ }{
    Solve $\theta_i$ by formula~(\ref{gmltrain}), and obtain the $i$th layer of data representation $H_i$;
}

\end{algorithm}

\subsection{Graph Regularizer Design}\label{graphdesignsection}

As mentioned above, the performance of date representation regularized with graph mainly lies on the design of the weight matrix $V$ since it encodes the local invariance information of the data space. In this section, we will focus on the weight matrix design of supervised learning and unsupervised learning in the context of data representation with auto-encoders.

\subsubsection{Unsupervised Learning}

In unsupervised learning, the label for the data is unavailable. We can only obtain the structure of the data from the local property of the data samples. There are three kinds of weights employed in our GAE, which are introduced as follows,
\begin{itemize}
  \item \textbf{KNN-graph:} It first constructs the $k$-nearest neighbor sets for each data sample. If $x_i$ lies in $x_j$'s $k$-nearest neighbor set, the weights $v_{ij}$ is set as the distance between these two data samples, that is $\exp(-\|x_i - x_j\|)$, otherwise $v_{ij}$ is set to zero.
  \item \textbf{$\epsilon$-graph:} It first constructs the $\epsilon$-neighbor sets for each data sample, the data sample $x_i$'s $\epsilon$-neighbor set contains all the data samples whose distances to $x_i$ are less than $\epsilon$. If $x_i$ lies in $x_j$'s $\epsilon$-nearest neighbor set, the weights $v_{ij}$ is set as the distance between these two data samples, that is $\exp(-\|x_i - x_j\|)$, otherwise $v_{ij}$ is set to zero.
  \item \textbf{$\mathfrak{l}_1$-graph:} The weight setting is considered to resolve the following optimization problem,
      \begin{equation}\label{l1opt}\nonumber
        v_{ij} = \arg \min \|x_i -\sum_{j=1\ldots n, j\neq i} v_{ij}x_j\| + \lambda \sum_{j=1\ldots n, j\neq i} |v_{ij}|
      \end{equation}

\end{itemize}

\subsubsection{Semi-supervised Learning}\label{semigraphsection}

In semi-supervised learning, the data labels are partially available, which brings some ground truth information to the estimation of the data representation. In our GAE, the graph regularizer design for semi-supervised learning task is similar to the unsupervised learning task. We first construct the $k$-nearest neighbor sets or $\epsilon$-neighbor sets for the whole data set, and set the weight $v_{ij}$ to zero if $x_i$ and $x_j$ are not neighbors. For the condition that $x_i$ and $x_j$ are neighbors, the weights are calculated as follow:
\begin{equation}\label{graphsemisupervised}
    v_{ij} = \left\{
                \begin{array}{cc}
                  \exp(-\|x_i-x_j\|) & \textrm{either $x_i$ or $x_j$ is unlabeled} \\
                       1 & \textrm{$x_i$, $x_j$ have the same label} \\
                       0 & \textrm{$x_i$, $x_j$ have different labels}
                \end{array}
                    \right.
\end{equation}

As mentioned before, the weights in the graph are computed by $\exp(-\|x_i - x_j\|)$, which is a value less than $1$ but larger than $0$. Since the labeled data will provide ground truth information, the weights of two samples with the same labels are directly set to $1$. And the weights between two samples with different labels are directly set to $0$.

The graph constraint constructed with formula~(\ref{graphsemisupervised}) is called semi-graph regularizer. Apparently, the marginal value $0$ and $1$ give the most confident level of the similarity since their corresponding are labeled. With this semi-graph regularizer, both labeled and unlabeled data samples are regarded fairly.

\section{Experimental Results}

In this section, comparison experiments are carried out to demonstrate the performance of our proposed method in tasks of both unsupervised learning and semi-supervised learning. To evaluate image representations learned by different methods quantitatively, the $k$-means clustering is applied to the representations learned by different methods. Two metrics, the normalized mutual information(MI) and accuracy(AC) are used to measure the clustering performance. For fair comparison, the dimension of learned representation through all algorithms are set to be the same.

The normalized mutual information(MI) is given in \cite{10.1109/TPAMI.2010.231}, which is a normalized measure to evaluate how similar two sets of clusters are. The accuracy (AC) is used to measure the percentage of correct labels compared to the ground truth label provided by the data set. Specifically, given a data sample $x_i$ with clustered label and ground truth label $c_i$ and $g_i$, the accuracy is defined as
\begin{equation}\label{acc}\nonumber
    AC=\frac{\sum_i \delta (g_i, map(c_i))}{n}
\end{equation}
where $n$ is total number of samples, $\delta(a,b)$ is delta function, which outputs $1$ when $a=b$ and outputs $0$ otherwise, $map(c)$ is the permutation mapping function that maps each clustered label $c_i$ to the best label provided by the data set. This function is implemented using the code published by \cite{10.1109/TPAMI.2010.231}. For the normalized mutual information, denote $C$ and $C'$ as the set of clusters obtained from the ground truth and our algorithm. We first compute the mutual information as follows,
\begin{equation}\label{mi}\nonumber
    \overline{MI}(C,C')=\sum_{c_i\in C, c'_j\in C'} p(c_i,c'_j)\log \frac{p(c_i,c'_j)}{p(c_i)p(c'_j)}
\end{equation}
where $p(c_i)$ and $p(c'_j)$ are the probabilities that a sample selected from the data set that belong to cluster $c_i$ and $c'_j$, $p(c_i,c'_j)$ is the probability that a sample selected from the data set that belong to both $c_i$ and $c'_j$. Then, the normalized mutual information can be computed as
\begin{equation}\label{nmi}\nonumber
    MI(C,C')=\frac{\overline{MI}(C,C')}{max(H(C),H(C'))}
\end{equation}
where $H(C)$ and $H(C')$ are the entropy of $C$ and $C'$. When $MI=1$, the two clusters are identical. when $MI=0$, the two clusters are independent.

The experimental results are all average value of multiple times of random experiment.
We set that the reduced representation of different dimension reduction techniques share the same dimensions.
The data sets employed for the experiments including: ORL\cite{samaria1994parameterisation}, Yale and COIL20, whose statistics are shown in table~\ref{data}.


\begin{table}[h]
\caption{Statistics of data sets employed in the experiments.}
\label{data}
\begin{center}
\begin{tabular}{cccc}
Datasets & \#Samples & \#Classes & \#Samples per class\\
\hline
ORL     & 400  &  40   &   10\\
Yale    & 165   &  15   &   11\\
COIL20  & 1440  &  20   &   72\\
\end{tabular}
\end{center}
\end{table}

\subsection{Variants Comparison}
Fine-tuning\cite{HinSal06} of a pre-trained deep network can sometimes improve the performance in many supervised learning tasks. However, in unsupervised learning, the weight matrix built on Euclidean distance may include some wrong information, i.e. samples with different labels may be connected. We can compute the error rate as the ratio of wrong connections and the total connections.

We construct a 2 layer auto-encoder, and implement layer-wise pre-training based on the method in section\ref{mlautoencoder}. Then we fine-tune the deep auto-encoder with the single graph regularization, i.e. only the second term of (\ref{gmltrain}). The input data is chosen from COIL20. It has 20 classes. In this experiment, we select 8 classes in random for comparison. So the unsupervised weight matrix is also kind of random based on the data set. Experiment result in table~\ref{Variantexperiment} shows that when the weight matrix is constructed with no error, the performance will be promoted with fine-tuning. However, when the weight matrix contains wrong connection, then the result turns out to be worse.

We also conduct an experiment that pre-train the GAE without reconstruction error but only with graph regularization which show in the last two rows in table~\ref{Variantexperiment}. The result is interesting that the deep architecture even cannot learn a meaningful representation. It may give some insights on the reconstruction error term.

As a result, in the next two sections, we train our deep auto-encoders using layer-wise pre-training with both reconstruction error and graph regularization.
\begin{table}[h]
\caption{Variant experiments.}
\label{Variantexperiment}
\begin{center}
\begin{tabular}{ccccc}
Model &Train Method &MI & AC & Error Rate\\
\hline
\multirow{2}*{GAE}      & Pre-train  &  0.8856   & 0.9167  & \multirow{2}*{3.1\%}\\
                        & Fine-tune  &  0.7666   & 0.6689  & \\
\multirow{2}*{GAE}      & Pre-train  &  0.8020   & 0.7465  & \multirow{2}*{0\%}\\
                        & Fine-tune  &  0.9026   & 0.8732  & \\
GAE                     & Pre-train  &  0.0156   & 0.2060  & \multirow{2}*{0\%}\\
No Reconstruction Error & Fine-tune  &  0.2129   & 0.0134  &\\
\end{tabular}
\end{center}
\end{table}

\subsection{Experiments in Unsupervised Learning}

For unsupervised learning task, all samples have no labels. So they are directly fed into the algorithm for measure evaluation. The dimension of the reduced representation is set to the number of classes in the input data set. The methods used for comparison including:
\begin{itemize}
  \item $k$-means: this is the baseline method which simply performs clustering in the original feature space.
  \item PCA: Principal Components Analysis. This is the classical linear technique for dimensional reduction.
  \item GAE: Graph Auto-Encoder (2 layers). It is a contribution of this paper that introduces the graph constraint to the auto-encoder. In the experiments, our GAE employs the KNN graph. There are two coefficients, $k$ for the number of neighbors in the KNN graph and $\lambda$ for the intensity of the graph regularizer. They are all selected by the grid based search.
  \item SAE: Sparse+Auto-encoder(2 layers)\cite{poultney2006efficient}. The sparse constraint is equipped to the auto-encoder, which is a very common constraint choice in the field of auto-encoder. The formula is given as follows
      \begin{equation}\label{sparseae}
        \hat{\theta} = \arg \min \|X-\hat{X}\|^2+\eta \sum_j KL(\rho | \rho_j)
      \end{equation}
      where $\rho$ is the user defined sparsity coefficient, $\rho_j$ is the average response of the $j$th hidden unit for the whole dataset. The penalty term can make the hidden response more sparse. The coefficient $\eta$ is also selected by grid based search.
  \item GNMF: Graph regularized Nonnegative Matrix Factorization. It is proposed in \cite{10.1109/TPAMI.2010.231}. It is a combination of nonnegative constraint and locally invariant constraint. The graph parameter settings are similar to the GAE, which employ the KNN graph and the coefficients $k$ and $\lambda$ are selected by the optimal grid search.
\end{itemize}

The results for whole datasets are shown in table~\ref{unsupervisedorl},~\ref{unsupervisedyale},~\ref{unsupervisedcoil20}. To randomize the experiments, we carry out evaluation with different cluster classes. For each given number of the cluster classes, we random choose the cluster classes from the whole datasets for $5$ times. One can see that graph regularized auto-encoder achieves the best performance. Although the GNMF and GAE are employed the same encodes on the locally invariant information, the auto-encoder which implement the nonlinear sigmoid scaling on the deep structure will performance better than the nonnegative matrix factorization based approach.


\begin{table*}[htb]
\caption{Results of the unsupervised learning tasks on ORL.}
\label{unsupervisedorl}
\begin{center}
\begin{tabular}{cccccccccc}
Class & 5 & 10 & 15 & 20 & 25 & 30 & 35 & 40 & Average \\
\hline
GAE MI & \textbf{0.8955} & \textbf{0.8967} & \textbf{0.8694} & \textbf{0.8531} & \textbf{0.8477} & \textbf{0.8315} & \textbf{0.8282} & \textbf{0.8344} & \textbf{0.8571}\\
SAE MI & 0.8632 &	0.8670& 	0.8438& 	0.8338& 	0.8423& 	0.8054& 	0.8062& 	0.7933& 0.8319\\
GNMF MI & 0.8464 &	0.7958& 	0.7634& 	0.7548& 	0.7759& 	0.7724& 	0.7627& 	0.7509& 0.7778\\
PCA MI & 0.8436 &	0.7969& 	0.7740& 	0.7492& 	0.7411& 	0.7354& 	0.7505& 	0.7500& 0.7676\\
Kmeans MI & 0.6921& 0.7392& 	0.6907& 	0.6939& 	0.6972& 	0.7088& 	0.7186& 	0.7078& 0.7060\\
\hline
GAE AC & \textbf{0.9333}& 	\textbf{0.8167}& 	0.7733& 	0.7483& 	\textbf{0.7400}& 	\textbf{0.6878}& 	\textbf{0.6876}& 	\textbf{0.6839}& 	\textbf{0.7589}\\
SAE AC & 0.9067& 	0.8067& 	\textbf{0.7800}& 	\textbf{0.7533}& 	0.7373& 	0.6867& 	0.6543& 	0.6217& 	0.7433\\
GNMF AC & 0.8840& 	0.7480& 	0.6773& 	0.6310& 	0.6272& 	0.6173& 	0.5714& 	0.5748& 	0.6664\\
PCA AC &0.8800& 	0.7380& 	0.6800& 	0.6190& 	0.5976& 	0.5567& 	0.5777& 	0.5625& 	0.6514\\
Kmeans AC & 0.7400& 0.6720& 	0.5933& 	0.5690& 	0.5552& 	0.5300& 	0.5354& 	0.5050& 	0.5875\\
\end{tabular}
\end{center}
\end{table*}

\begin{table*}[htb]
\caption{Results of the unsupervised learning tasks on Yale.}
\label{unsupervisedyale}
\begin{center}
\begin{tabular}{ccccccc}
Class & 3 & 6 & 9 & 12 & 15 &Average \\
\hline
GAE MI & \textbf{0.7283}& 	\textbf{0.5748}& 	\textbf{0.5467}& 	\textbf{0.5395}& 	\textbf{0.5689}& 	\textbf{0.5916}\\
SAE MI & 0.6140& 	0.5154& 	0.5386& 	0.5171& 	0.5350& 	0.5440\\
GNMF MI & 0.5175& 	0.5111& 	0.5130& 	0.5034& 	0.4518& 	0.4994\\
PCA MI & 0.3650& 	0.3883& 	0.4686& 	0.4859& 	0.5103& 	0.4436\\
Kmeans MI & 0.4198&	0.3082& 	0.4035& 	0.4251& 	0.4532& 	0.4020\\
\hline
GAE AC & \textbf{0.8889}& 	\textbf{0.6515}& 	0.5623& 	\textbf{0.5177}& 	\textbf{0.5313}& 	\textbf{0.6303}\\
SAE AC & 0.8485& 	0.6313& 	\textbf{0.5993}& 	0.4975& 	0.4828& 	0.6119\\
GNMF AC & 0.7758& 	0.5818& 	0.5414& 	0.4818& 	0.4162& 	0.5594\\
PCA AC  &0.6424 &	0.4909& 	0.4970& 	0.4652& 	0.4412& 	0.5073\\
Kmeans AC &0.6303& 	0.4303& 	0.4222& 	0.4182& 	0.4048& 	0.4612\\
\end{tabular}
\end{center}
\end{table*}

\begin{table*}[htb]
\caption{Results of the unsupervised learning tasks on COIL20.}
\label{unsupervisedcoil20}
\begin{center}
\begin{tabular}{ccccccccc}
Class & 6 & 8 & 10 & 12 & 14 & 16 & 20 & Average \\
\hline
GAE MI & \textbf{0.9739}& 	0.8888& 	\textbf{0.8762}& 	\textbf{0.8684}& 	\textbf{0.8657}& 	\textbf{0.8552}& 	\textbf{0.8502}& 	\textbf{0.8826}\\
SAE MI & 0.9264 &	\textbf{0.8953}& 	0.8527& 	0.8280& 	0.8344& 	0.8078& 	0.8013& 	0.8494\\
GNMF MI & 0.8690 &	0.8469& 	0.8696& 	0.8302& 	0.8379& 	0.8460& 	0.8449& 	0.8492\\
PCA MI & 0.7350 &	0.7544& 	0.7790& 	0.7988& 	0.7686& 	0.7894& 	0.7817& 	0.7724\\
Kmeans MI & 0.6550& 0.7490& 	0.7714& 	0.7720& 	0.7382& 	0.7392& 	0.7354& 	0.7372\\
\hline
GAE AC &\textbf{0.9846}& 	\textbf{0.8779}& 	\textbf{0.8495}& 	\textbf{0.8279}& 	\textbf{0.8056}& 	\textbf{0.7888}& 	\textbf{0.7981}& 	\textbf{0.8475}\\
SAE AC &0.9483& 	0.8709& 	0.8352& 	0.7731& 	0.7665& 	0.7237& 	0.7255& 	0.8062\\
GNMF AC &0.8690& 	0.8097& 	0.8358& 	0.7692& 	0.7579& 	0.7566& 	0.7172& 	0.7879\\
PCA AC  &0.7435& 	0.7035& 	0.7553& 	0.7461& 	0.6917& 	0.7148& 	0.6871& 	0.7203\\
Kmeans AC &0.7106& 	0.7372& 	0.7472& 	0.7338& 	0.6579& 	0.6646& 	0.6096& 	0.6944\\
\end{tabular}
\end{center}
\end{table*}


\subsection{Experiments in Semi-supervised Learning}

\begin{table*}[htb]
\caption{Results of the semi-supervised learning tasks for ORL.}
\label{semisupervisedorl}
\begin{center}
\begin{tabular}{cccccccccc}
Class & 5 & 10 & 15 & 20 & 25 & 30 & 35 & 40 & Average \\
\hline
GAE MI & \textbf{0.9242} &	\textbf{0.9335}& 	\textbf{0.8996}& 	\textbf{0.8872}& 	\textbf{0.8772}& 	\textbf{0.8787}& 	\textbf{0.8681}& 	\textbf{0.8694}& 	\textbf{0.8923}\\
CNMF MI & 0.8257 &	0.7998& 	0.8275& 	0.8161& 	0.8007& 	0.7943& 	0.7885& 	0.7873& 	0.8050\\
\hline
GAE AC & \textbf{0.9533}& 	\textbf{0.9167}& 	\textbf{0.8400}& 	\textbf{0.7783}& 	\textbf{0.7800}& 	\textbf{0.7756}& 	\textbf{0.7543}& 	\textbf{0.7325}& 	\textbf{0.8163}\\
CNMF AC & 0.8680& 	0.7580& 	0.7520& 	0.7290& 	0.6824& 	0.6620& 	0.6354& 	0.6250& 	0.7140\\
\end{tabular}
\end{center}
\end{table*}

\begin{table*}[htb]
\caption{Results of the semi-supervised learning tasks for Yale.}
\label{semisupervisedyale}
\begin{center}
\begin{tabular}{ccccccc}
Class & 3 & 6 & 9 & 12 & 15 &Average \\
\hline
GAE MI & \textbf{0.6400}& 	\textbf{0.7322}& 	\textbf{0.6336}& 	\textbf{0.6126}& 	\textbf{0.6550}& 	\textbf{0.6547}\\
CNMF MI & 0.5915& 	0.5733& 	0.5620& 	0.5710& 	0.5543& 	0.5704\\
\hline
GAE AC & \textbf{0.8384}& 	\textbf{0.7172}& 	\textbf{0.6604}& 	\textbf{0.6162}& 	\textbf{0.5818}& 	\textbf{0.6828}\\
CNMF AC & 0.7636& 	0.6636& 	0.5939& 	0.5439& 	0.4949& 	0.6120\\
\end{tabular}
\end{center}
\end{table*}

\begin{table*}[htb]
\caption{Results of the semi-supervised learning tasks for COIL20.}
\label{semisupervisedcoil20}
\begin{center}
\begin{tabular}{ccccccccc}
Class & 6 & 8 & 10 & 12 & 14 & 16 & 20 & Average \\
\hline
GAE MI & \textbf{0.9182}& 	\textbf{0.9406}& 	\textbf{0.9281}& 	\textbf{0.9131}& 	\textbf{0.9215}& 	\textbf{0.8908}& 	\textbf{0.8642}& 	\textbf{0.9109}\\
CNMF MI & 0.7861& 	0.8129& 	0.7580& 	0.7606& 	0.7576& 	0.7904& 	0.7440& 	0.7728\\
\hline
GAE AC &\textbf{0.9529}& 	\textbf{0.9630}& 	\textbf{0.9000}& 	\textbf{0.9028}& 	\textbf{0.8991}& 	\textbf{0.8313}& 	\textbf{0.8236}& 	\textbf{0.8961}\\
CNMF AC &0.7755& 	0.7951& 	0.7396& 	0.7043& 	0.7131& 	0.7109& 	0.6505& 	0.7270\\
\end{tabular}
\end{center}
\end{table*}


For semi-supervised learning task, a small part of the samples are labeled. In this experiment, these labeled samples are selected in random. For COIL20, $10\%$ samples are labeled in each class, so there are 7 labeled samples in each class. For ORL and Yale, $20\%$ are labeled, then $2$ samples are labeled in each class~\footnote{Here one labeled sample is meaningless to both CNMF and SGAE, so we label $20\%$ of the samples in each class.}. Still referring to the unsupervised learning experiments, the dimension of the learned representation is equal to the number of classes in the data set. The comparison methods used in this experiments including:
\begin{itemize}
  \item CNMF: constrained NMF. It is proposed in \cite{10.1109/TPAMI.2011.217}. In their framework, the samples with the same label are required to have the same representation in the reduced space. There is no user defined parameters either.
  \item SGAE: Semi-Graph regularized Auto-Encoder (2 layers). It is a representation learning algorithm proposed in this paper consisting of the auto-encoder regularized by the semi-graph regularizer presented in section~\ref{semigraphsection}. The parameters include the intensity of the graph constraint, $\lambda$ and the number of neighbors in the KNN graph, $k$. Similarly to the GAE in unsupervised learning experiment, these two parameters are selected by optimal grid search.
\end{itemize}

The clustering results on all classes of the datasets are shown in table~\ref{semisupervisedorl},~\ref{semisupervisedyale},~\ref{semisupervisedcoil20}. Similarly to the randomize experiment of the unsupervised learning, we also conduct the randomize experiment for semi-supervised learning on different number of classes. The classes for the experiment are also randomly sampled from the whole datasets with $5$ times, and the average clustering results are shown in rightmost column of the table. It can be found that the semi-graph regularized auto-encoder gives a significant improvement of performance compared to the constrained NMF. The reason may be that the CNMF only utilizes the labeled data while ignoring the geometric structure hidden in the unlabeled data. When it comes to the semi-graph regularized auto-encoder, all the information from labeled and unlabeled data are all considered. As we expected, semi-graph regularized auto-encoder achieves better performance compared to the unsupervised clustering results in table~\ref{unsupervisedorl},~\ref{unsupervisedyale},~\ref{unsupervisedcoil20}.

\section{Conclusion}
In this paper, we proposed a novel graph regularized auto-encoder, which can learn a locally invariant representation of the images for both unsupervised and semi-supervised learning tasks. In unsupervised learning, our approach trains the image representation by an multiple-layer auto-encoder regularized with the graph, which encodes the locally neighborhood relationships of the original data. And in semi-supervised learning, the graph regularizer used in our auto-encoders is extended to semi-graph regularizer, which adds the penalty and reward obtained from the labeled data points to the locally neighborhood weight matrix. Experimental results on image clustering show our method provides better performance comparing with the stat-of-the-art approaches.
The further work may focus on investigating the affections of the parameter settings in GAE and the impacts of the deep structure is also a possible future work.


{\small
\bibliographystyle{ieee}
\bibliography{egbib}

\begin{thebibliography}{10}\itemsep=-1pt

\bibitem{DBLP:conf/nips/BelkinN01}
M.~Belkin and P.~Niyogi.
\newblock Laplacian eigenmaps and spectral techniques for embedding and
  clustering.
\newblock In {\em NIPS}, pages 585--591, 2001.

\bibitem{MR_JMLR_06}
M.~Belkin, P.~Niyogi, and V.~Sindhwani.
\newblock Manifold regularization: A geometric framework for learning from
  labeled and unlabeled examples.
\newblock {\em Journal of Machine Learning Research}, 7:2399--2434, 2006.

\bibitem{10.1109/TPAMI.2013.50}
Y.~Bengio, A.~Courville, and P.~Vincent.
\newblock Representation learning: A review and new perspectives.
\newblock {\em IEEE Transactions on Pattern Analysis and Machine Intelligence},
  35(8):1798--1828, 2013.

\bibitem{10.1109/TPAMI.2010.231}
D.~Cai, X.~He, J.~Han, and T.~S. Huang.
\newblock Graph regularized nonnegative matrix factorization for data
  representation.
\newblock {\em IEEE Transactions on Pattern Analysis and Machine Intelligence},
  33(8):1548--1560, 2011.

\bibitem{Goodfellow+etal09:invariance}
I.~Goodfellow, Q.~Le, A.~Saxe, H.~Lee, and A.~Y. Ng.
\newblock Measuring invariances in deep networks.
\newblock In {\em Advances in Neural Information Processing Systems 22}, pages
  646--654. 2009.

\bibitem{Hadsell:2006:DRL:1153171.1153654}
R.~Hadsell, S.~Chopra, and Y.~LeCun.
\newblock Dimensionality reduction by learning an invariant mapping.
\newblock In {\em Proceedings of the 2006 IEEE Computer Society Conference on
  Computer Vision and Pattern Recognition - Volume 2}, CVPR '06, pages
  1735--1742, Washington, DC, USA, 2006. IEEE Computer Society.

\bibitem{HinSal06}
G.~Hinton and R.~Salakhutdinov.
\newblock Reducing the dimensionality of data with neural networks.
\newblock {\em Science}, 313(5786):504 -- 507, 2006.

\bibitem{DBLP:conf/nips/HintonZ93}
G.~E. Hinton and R.~S. Zemel.
\newblock Autoencoders, minimum description length and helmholtz free energy.
\newblock In {\em NIPS}, pages 3--10, 1993.

\bibitem{NIPS2012_0534}
A.~Krizhevsky, I.~Sutskever, and G.~Hinton.
\newblock Imagenet classification with deep convolutional neural networks.
\newblock In P.~Bartlett, F.~Pereira, C.~Burges, L.~Bottou, and K.~Weinberger,
  editors, {\em Advances in Neural Information Processing Systems 25}, pages
  1106--1114. 2012.

\bibitem{DBLP:conf/icml/LarochelleB08}
H.~Larochelle and Y.~Bengio.
\newblock Classification using discriminative restricted boltzmann machines.
\newblock In {\em ICML}, pages 536--543, 2008.

\bibitem{DBLP:conf/nips/LeeEN07}
H.~Lee, C.~Ekanadham, and A.~Y. Ng.
\newblock Sparse deep belief net model for visual area v2.
\newblock In {\em NIPS}, 2007.

\bibitem{10.1109/TPAMI.2011.217}
H.~Liu, Z.~Wu, D.~Cai, and T.~S. Huang.
\newblock Constrained nonnegative matrix factorization for image
  representation.
\newblock {\em IEEE Transactions on Pattern Analysis and Machine Intelligence},
  34(7):1299--1311, 2012.

\bibitem{min2010deep}
M.~R. Min, L.~Maaten, Z.~Yuan, A.~J. Bonner, and Z.~Zhang.
\newblock Deep supervised t-distributed embedding.
\newblock In {\em Proceedings of the 27th International Conference on Machine
  Learning (ICML-10)}, pages 791--798, 2010.

\bibitem{mobahi2009deep}
H.~Mobahi, R.~Collobert, and J.~Weston.
\newblock Deep learning from temporal coherence in video.
\newblock In {\em Proceedings of the 26th Annual International Conference on
  Machine Learning}, pages 737--744. ACM, 2009.

\bibitem{poultney2006efficient}
C.~Poultney, S.~Chopra, Y.~L. Cun, et~al.
\newblock Efficient learning of sparse representations with an energy-based
  model.
\newblock In {\em Advances in neural information processing systems}, pages
  1137--1144, 2006.

\bibitem{NIPS2007_1118}
M.~Ranzato, Y.-L. Boureau, and Y.~LeCun.
\newblock Sparse feature learning for deep belief networks.
\newblock In J.~Platt, D.~Koller, Y.~Singer, and S.~Roweis, editors, {\em
  Advances in Neural Information Processing Systems 20}, pages 1185--1192. MIT
  Press, Cambridge, MA, 2008.

\bibitem{DBLP:conf/nips/RanzatoPCL06}
M.~Ranzato, C.~S. Poultney, S.~Chopra, and Y.~LeCun.
\newblock Efficient learning of sparse representations with an energy-based
  model.
\newblock In {\em NIPS}, pages 1137--1144, 2006.

\bibitem{Rifai:2011:HOC:2034117.2034159}
S.~Rifai, G.~Mesnil, P.~Vincent, X.~Muller, Y.~Bengio, Y.~Dauphin, and
  X.~Glorot.
\newblock Higher order contractive auto-encoder.
\newblock In {\em Proceedings of the 2011 European conference on Machine
  learning and knowledge discovery in databases - Volume Part II}, ECML
  PKDD'11, pages 645--660, Berlin, Heidelberg, 2011. Springer-Verlag.

\bibitem{citeulike:9426230}
S.~Rifai, P.~Vincent, X.~Muller, X.~Glorot, and Y.~Bengio.
\newblock Contractive {Auto-Encoders}: Explicit invariance during feature
  extraction.
\newblock In {\em ICML}, 2011.

\bibitem{Roweis2000}
S.~T. Roweis and L.~K. Saul.
\newblock {Nonlinear Dimensionality Reduction by Locally Linear Embedding}.
\newblock {\em Science}, 290(5500):2323--2326, 2000.

\bibitem{samaria1994parameterisation}
F.~S. Samaria and A.~C. Harter.
\newblock Parameterisation of a stochastic model for human face identification.
\newblock In {\em Applications of Computer Vision, 1994., Proceedings of the
  Second IEEE Workshop on}, pages 138--142. IEEE, 1994.

\bibitem{DBLP:conf/nips/SocherHBMN12}
R.~Socher, B.~Huval, B.~P. Bath, C.~D. Manning, and A.~Y. Ng.
\newblock Convolutional-recursive deep learning for 3d object classification.
\newblock In {\em NIPS}, pages 665--673, 2012.

\bibitem{Tenenbaum2000global}
J.~B. Tenenbaum, V.~de~Silva, and J.~C. Langford.
\newblock A global geometric framework for nonlinear dimensionality reduction.
\newblock {\em Science}, 290(5500):2319--23, Dec 2000.

\bibitem{Vincent:2011:CSM:2000609.2000610}
P.~Vincent.
\newblock A connection between score matching and denoising autoencoders.
\newblock {\em Neural Comput.}, 23(7):1661--1674, July 2011.

\bibitem{Vincent:2008:ECR:1390156.1390294}
P.~Vincent, H.~Larochelle, Y.~Bengio, and P.-A. Manzagol.
\newblock Extracting and composing robust features with denoising autoencoders.
\newblock In {\em Proceedings of the 25th international conference on Machine
  learning}, ICML '08, pages 1096--1103, New York, NY, USA, 2008. ACM.

\bibitem{Vincent:2010:SDA:1756006.1953039}
P.~Vincent, H.~Larochelle, I.~Lajoie, Y.~Bengio, and P.-A. Manzagol.
\newblock Stacked denoising autoencoders: Learning useful representations in a
  deep network with a local denoising criterion.
\newblock {\em J. Mach. Learn. Res.}, 11:3371--3408, Dec. 2010.

\end{thebibliography}
}

\end{document}